\documentclass[10pt,twocolumn,letterpaper]{article}

\usepackage{cvpr}              

\usepackage[dvipsnames]{xcolor}

\definecolor{cvprblue}{rgb}{0.21,0.49,0.74}
\usepackage[pagebackref,breaklinks,colorlinks,citecolor=cvprblue]{hyperref}
\usepackage{nicematrix}
\usepackage{multirow}

\newcommand{\zss}[1]{\textcolor{black}{#1}}
\def\paperID{14533} 
\def\confName{CVPR}
\def\confYear{2024}
\makeatletter
\def\thanks#1{\protected@xdef\@thanks{\@thanks
        \protect\footnotetext{#1}}}
\makeatother

\title{Local-consistent Transformation Learning for Rotation-invariant \\
Point Cloud Analysis}

\begin{document}
\author{
    Yiyang~Chen$^{1}$\thanks{\dag Corresponding authors} \quad
    Lunhao~Duan$^{2}$ \quad
    Shanshan~Zhao$^{3\dag}$ \quad
    Changxing~Ding$^{1, 4\dag}$ \quad
    Dacheng~Tao$^{5}$ \\
    $^{1}$South~China~University~of~Technology \quad $^{2}$Wuhan University \quad $^{3}$The University of Sydney \\ \quad $^{4}$Pazhou~Lab \quad $^{5}$Nanyang Technological University\\
    {\tt\small eeyychen@mail.scut.edu.cn, lhduan@whu.edu.cn}\\
    {\tt\small \{sshan.zhao00, dacheng.tao\}@gmail.com, chxding@scut.edu.cn}
}

\maketitle
\begin{abstract}
    
Rotation invariance is an important requirement for point shape analysis. To achieve this, current state-of-the-art methods attempt to construct the local rotation-invariant representation through learning or defining the local reference frame (LRF). Although efficient, these LRF-based methods suffer from perturbation of local geometric relations, resulting in suboptimal local rotation invariance.
To alleviate this issue, we propose a Local-consistent Transformation (LocoTrans) learning strategy. Specifically, we first construct the local-consistent reference frame (LCRF) by considering the symmetry of the two axes in LRF. In comparison with previous LRFs, our LCRF is able to preserve local geometric relationships better through performing local-consistent transformation. However, as the consistency only exists in local regions, the relative pose information is still lost in the intermediate layers of the network. We mitigate such a relative pose issue by developing a relative pose recovery (RPR) module. RPR aims to restore the relative pose between adjacent transformed patches. Equipped with LCRF and RPR, our LocoTrans is capable of learning local-consistent transformation and preserving local geometry, which benefits rotation invariance learning. Competitive performance under arbitrary rotations on both shape classification and part segmentation tasks and ablations can demonstrate the effectiveness of our method. Code will be available publicly at \url{https://github.com/wdttt/LocoTrans}.

\begin{figure}[htbp]
  \centering
  \includegraphics[width=\linewidth]{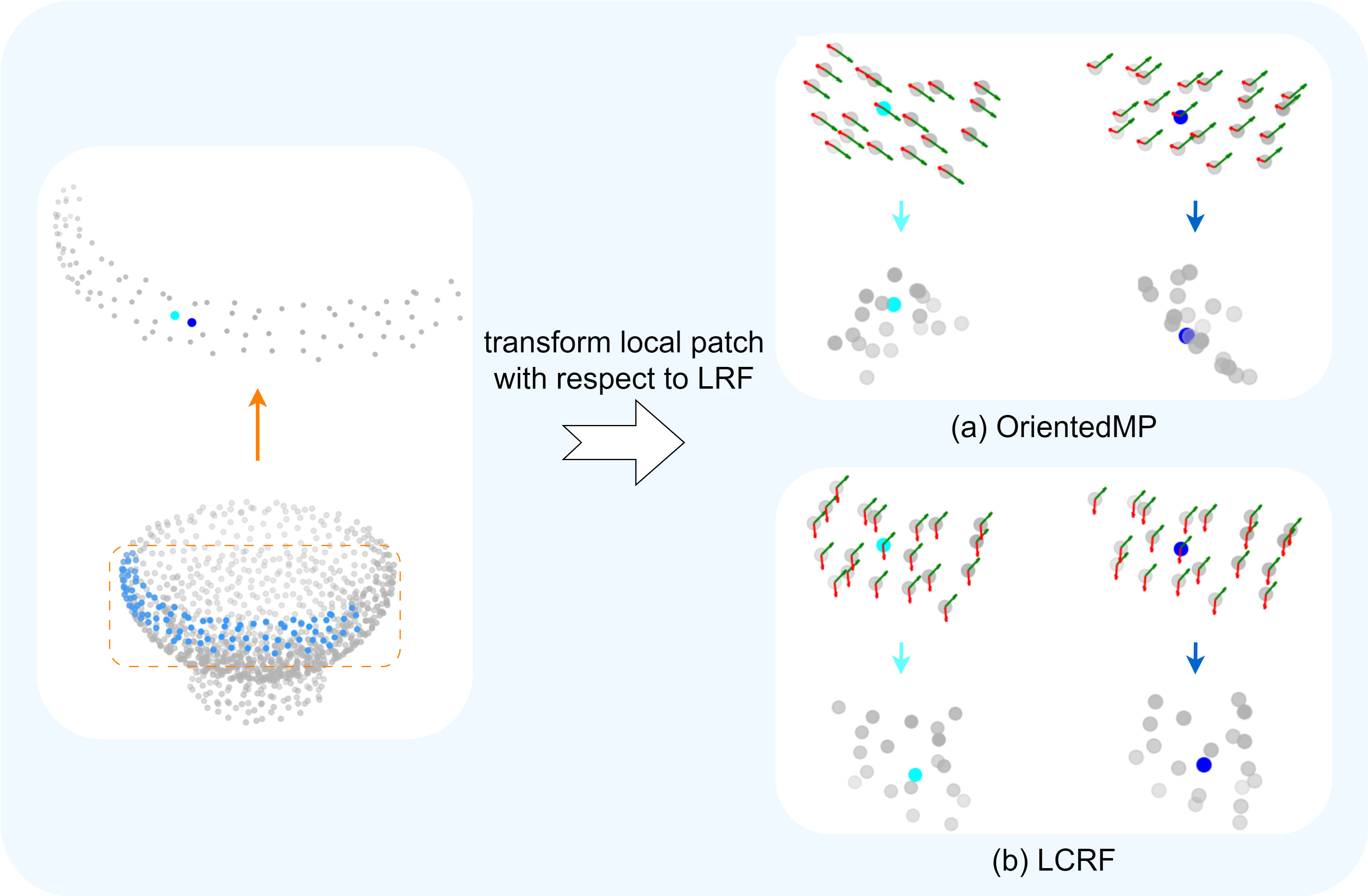} 
      \caption{Illustration of LRFs and the points transformation. For two reference points marked by Blue and Cyan, we use the LRF in OrientedMP~\cite{Luo_2022_CVPR} and our proposed LCRF to transform the local patches centered in them, respectively. By observing (a) and (b), we can find that our LCRF is able to yield local-consistent transformation and thus preserve the local shapes better. In comparison, OrientedMP only generates consistent orientation in one axis (marked by Red) but fails in the other (marked by Green).
      }
      \vspace{-10pt}
  \label{fig:struture} 
\end{figure}
\end{abstract}
\section{Introduction}
    Point cloud analysis~\cite{Qi_2017_CVPR,Zhao_2021_ICCV, Afham_2022_CVPR, Zhang_2022_CVPR, yang2023geometry, jin2023context, duan2023condaformer} is a fundamental task in 3D vision. Existing works~\cite{NIPS2017_d8bf84be,dgcnn} have achieved excellent performance in classification and part segmentation \zss{for pre-aligned point cloud data}. 
 \zss{However, these works tend to suffer severe performance drops when processing 3D objects with arbitrary rotations, as they do not learn the rotation-invariant feature representations. The lack of the capability for rotation invariance limits the application of the point cloud analysis model in practical situations. Currently, many attempts~\cite{Chen_2019_CVPR,Luo_2022_CVPR,NEURIPS2020_5d0cb12f,Xu_2021_ICCV,zhang2020learning} have been made to improve the performance under arbitrary rotations.}

\zss{To achieve rotation invariance, previous works~\cite{zhang-riconv-3dv19,Chen_2019_CVPR,Xu_2021_ICCV} propose to exploit local relative geometric relationships that remain unchanged under rotations, such as distance and angle, for feature learning, instead of the 3D coordinates. However, such a strategy does not fully use 3D geometric information. Recently, some methods~\cite{zhang2020learning,Luo_2022_CVPR,Chen_2022_CVPR} aim to learn rotation-invariant representations by building local reference frames (LRFs). These LRF-based methods construct three orthonormal basis vectors for each point as the reference frame and then project the original local points into the frame.
The orientations of LRF can be defined from geometric information, such as local barycenter, global center, and normal vector, or learned from input point cloud data. For instance, OrientedMP~\cite{Luo_2022_CVPR} is a pioneering work that employs the equivariant network to extract local features for building LRF. }

However, constructing local rotation-invariant representation via existing LRFs inevitably changes the original local geometry. 
\zss{As shown in Fig.~\ref{fig:struture}(a), for two adjacent points (marked by Cyan and Blue), we transform the local regions centered by them using the LRF built by OrientedMP, and different orientations are observed. As a result, the network might be struggling to learn identical representations for them due to the loss of relative pose.}
Moreover, this perturbation may lead to an ambiguous representation when applying vanilla convolution directly to points that have lost their relative pose and been assigned within a local patch after a k-nearest-neighbors (KNN) operation,
as observed in existing works~\cite{zhang2020learning,Chen_2022_CVPR}. 
\zss{To analyze the reason causing such an issue,  we further visualize the learned orientations along two axes in OrientedMP. As illustrated in Fig.~\ref{fig:struture}(a), the local consistency only exists in one orientation but not in the other. Therefore,  
we attribute the perturbation of geometry relationships to the local inconsistency of learned LRF.}

\zss{Motivated by the analysis above, we propose \textit{\textbf{Lo}cal-\textbf{co}nsistent \textbf{Trans}formation} (\textbf{LocoTrans}) learning strategy, aiming to learn local-consistent reference frame and achieving better local geometry preservation.}
\zss{Specifically, following OrientedMP, we also exploit the equivariant network for constructing LRF, as the equivariant information can be leveraged to eliminate rotations existing in data.
OrientedMP utilizes this property to construct two axes of LRF based on Gram-Schmidt orthogonalization with equivariant features. However, it suffers from a learning imbalance between the two axes due to the asymmetry of their definition.
Therefore, to address this issue, 
we select two axes of LRF that satisfy the symmetry based on the angular bisector of the equivariant features. Then two constraints including orthogonality and consistency are employed to optimize the LRF, yielding local-consistent reference frame (LCRF).
Thus, LCRF is able to achieve local-consistent transformation during the construction of local rotation-invariant representation, mitigating disturbances in local geometry. As shown in Fig.~\ref{fig:struture}(b), our LCRF achieves consistent orientations in both two axes for the two neighboured points. Nevertheless, 
since we can only achieve consistency in a local region, in the intermediate layers of the network, the relative pose between adjacent points is still changed, which is called the relative pose issue~\cite{Chen_2022_CVPR,zhang2020learning}. We alleviate this issue by developing a relative pose recovery (RPR) module. In our RPR, we
utilize the equivariant features that contain original global information to improve the learned rotation-invariant features.
}

Our LocoTrans is implemented through a fusion network comprising equivariant and invariant branches, with the invariant branch relying on the equivariant one to achieve LCRF and RPR.
The key contributions of this paper can be summarized as follows:
\begin{itemize}
  \item 
  We propose LocoTrans to achieve rotation-invariant feature learning while keeping local geometric information.
  
  \item We develop LCRF by 
  a new definition for axes of LRF satisfying the symmetry, 
  which is able to learn local-consistent reference frame; We also design an RPR module to address the relative pose issue.

  \item We conduct experiments to evaluate whether our network is rotation-invariant to arbitrary rotations in both classification and part segmentation tasks. The results show the effectiveness of our method.
\end{itemize}
\section{Related Works}
    \subsection{Point Cloud Shape Analysis}
Methods for 3D point cloud shape analysis can be categorized into three primary groups: projection-based~\cite{Qi_2016_CVPR,Su_2015_ICCV}, voxel-based~\cite{maturana2015voxnet,Wu_2015_CVPR}, and point-based~\cite{Qi_2017_CVPR,NIPS2017_d8bf84be,dgcnn}. The first two categories are constrained by geometric information loss and high computational costs, respectively, which currently have been explored by some works~\cite{Klokov_2017_ICCV, 3DSemanticSegmentationWithSubmanifoldSparseConvNet}. In contrast, point-based methods 
work on point clouds without additional operations, with PointNet~\cite{Qi_2017_CVPR} pioneering direct processing of point sets. 
Subsequent works consider local context and prioritize various aspects, such as abstracting strategies~\cite{NIPS2017_d8bf84be,Zhao_2019_CVPR}, convolution operations~\cite{Xu_2021_CVPR,Zhou_2021_ICCV}, and feature aggregation~\cite{Xiang_2021_ICCV,chen2022discard}.
However, a common drawback in many existing methods is a lack of robustness to rotations, leading to a significant performance decline when confronted with arbitrarily rotated data.
In our work, we address this limitation by introducing rotation invariance to the 3D network by exploring the construction of local reference frames.
\subsection{Rotation Invariance Methods}
Recently, some works~\cite{yu2020deep,xiao2020endowing,Li_2021_ICCV} address the challenge of rotations by converting rotated point clouds to canonical poses through principal component analysis (PCA) to obtain rotation-invariant features. \citet{Li_2021_ICCV} comprehensively summarize ambiguities in canonical poses and mitigate their impact through blending strategies.
Despite these efforts, complete elimination of ambiguities remains challenging, necessitating extensive augmentation for approximate invariance.
Additionally, constructing rotation-invariant features within local patches has been widely explored, involving handcrafted features with local geometry replacing Cartesian coordinates as input. 
RI-CNN~\cite{zhang-riconv-3dv19} designs a four-dimensional feature with distance and angle
while PaRI-Conv~\cite{Chen_2022_CVPR} utilizes the normal vector to extend it to eight dimensions. 
In the absence of normal vectors, SGMNet~\cite{Xu_2021_ICCV} connects every pair of points in a local neighborhood for a full geometrical description.
LRF-based methods~\cite{zhang2020learning,Chen_2022_CVPR,Luo_2022_CVPR,NEURIPS2020_5d0cb12f} introduce rotation invariance through building local reference frames, which can be used to transform local coordinates.
However, these methods often sacrifice the original pose information between local patches, resulting in performance limitations. To address this issue, some methods~\cite{zhang2020learning,Chen_2022_CVPR} attempt to restore relative pose with hand-crafted features.
In this paper, we first alleviate relative pose changes with local-consistent reference frame and then leverage equivariant network to capture local geometry for restoring relative pose.

\subsection{Rotation Equivariance Methods}
Currently, typical solutions to equivariance can be divided into tensor field-based and vector-based.
Tensor field-based methods~\cite{thomas2018tensor,NEURIPS2020_15231a7c,Poulenard_2021_CVPR} constrain convolutional kernels to adhere to the spherical harmonics family to achieve equivariance but are limited by significant memory requirements. In comparison,
vector-based methods~\cite{Deng_2021_ICCV,jing2021learning,satorras2021n,schutt2021equivariant,su2022svnet} transform standard neural network representations from scalar to vector to achieve equivariance. 
Typically, 
activation functions are often used to introduce non-linearity into the network for better learning.
However, maintaining equivariance in vector-based networks requires linear layer compositions in encoders, limiting network performance. 
To address this limitation, some works~\cite{Deng_2021_ICCV,jing2021learning} explore introducing non-linearity through a combination of linear layers. 
VNN~\cite{Deng_2021_ICCV} achieves this through vector truncation operations, while GVP~\cite{jing2021learning} scales vectors based on their norms.
In our paper, we leverage features output by an equivariant network to enhance the learning of local rotation-invariant representation.

\section{Method}
    In this paper, our objective is to learn local rotation-invariant representation with local-consistent reference frames. 
In the following, we first introduce the overview of our method in Sec.~\ref{sec:over}, then detail our Local-consistent Transformation  (LocoTrans) learning in Sec.~\ref{sec:LocoTrans} by outlining the components of the local-consistent reference frame (LCRF) construction and the relative pose recovery (RPR) module.

\subsection{Overview}
\label{sec:over}
\begin{figure*}[htbp]
  \centering
  \includegraphics[width=0.9\linewidth]{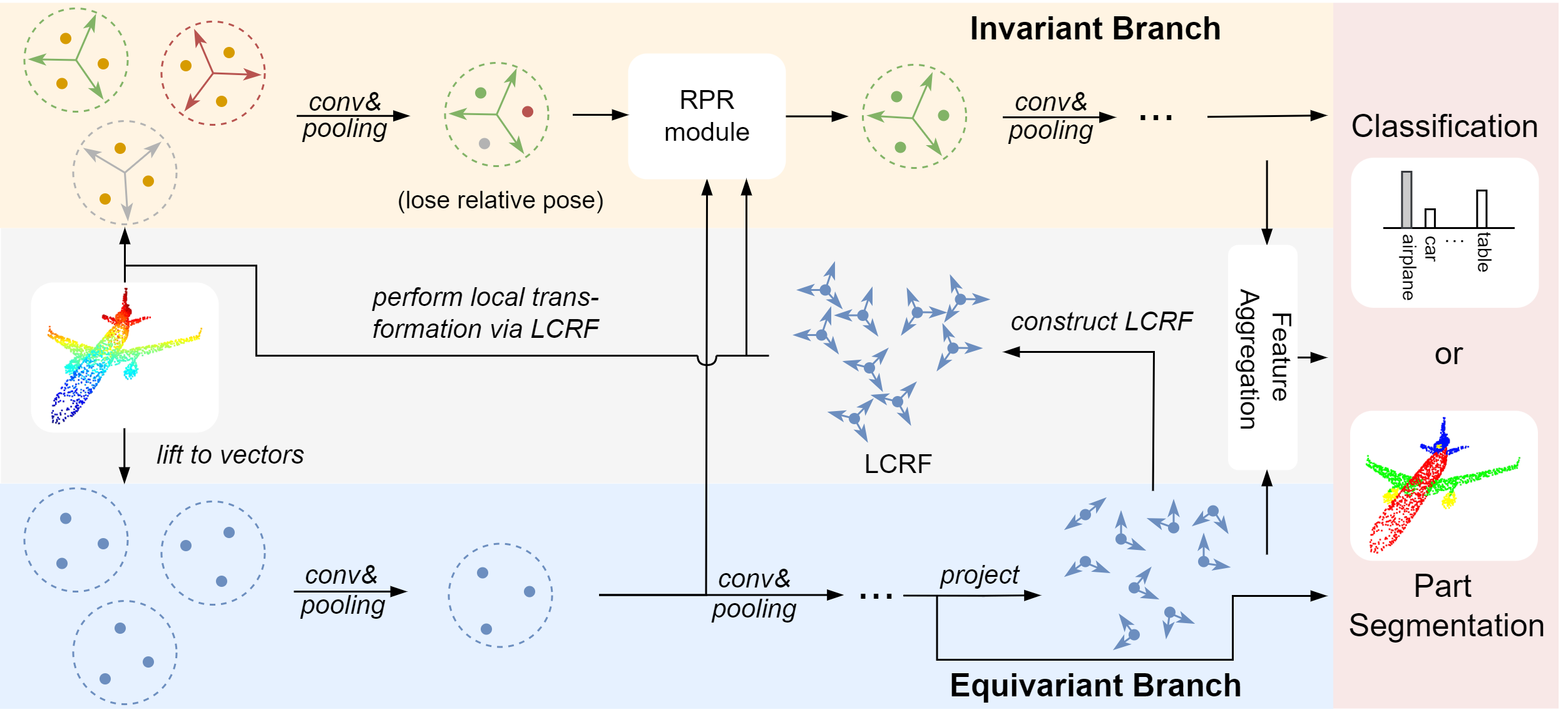} 
    \vspace{-5pt}
      \caption{Overall framework of our LocoTrans. 
      Our network consists of the invariant branch and the equivariant branch. 
      To reduce the local geometric perturbation caused by local region transformation, we construct LCRF using the features from the equivariant branch to perform the local-consistent transformation. However, such a perturbation is difficult to completely eliminate. 
      The RPR module is presented to alleviate this issue, restoring the relative pose between local patches. At the output level, we merge the features from two branches to aggregate information.
      } 
      \vspace{-15pt}
  \label{fig:framwork} 
\end{figure*}

The overall framework of our approach is shown in Fig.~\ref{fig:framwork}.
Our LocoTrans achieves LCRF and RPR with equivariant features. Thus, in our framework, we introduce a fusion network comprising two branches: the invariant branch and the equivariant branch. Given the significant success of DGCNN~\cite{dgcnn} in the field of point cloud shape analysis, we adopt it as the backbone for both branches. 
Our fusion network takes raw point clouds as input. 
Initially, we feed the raw points into the equivariant branch to get equivariant features that are then used to construct LCRF and used in RPR module of the invariant branch.
At the output stage, the attention mechanism, widely applied in feature fusion~\cite{Zhuang_2021_ICCV,yan20222dpass} to aggregate information, is employed to fuse features from the invariant branch and the equivariant branch.

Before delving into the details of our method, we establish some notations. Given a point cloud $P=[p_1,p_2,...,p_N]\in\mathbb{R}^{N\times 3}$ and an arbitrary rotation $R\in$ SO(3), 
the rotation-invariant representation $\delta(p)$ satisfies:
\begin{equation}
    \begin{aligned}
        \delta(Rp) = \delta(p),
    \end{aligned}
\end{equation}
where $p\in P$ and $\delta(\cdot)$ is the function to build 
rotation-invariant representation.
The  equivariant features we require in our LocoTrans are generated from an equivariant network $h$ which satisfies:
\begin{equation}
    \begin{aligned}
        h(Rp) = Rh(p) = Rv,
    \end{aligned}
    \label{eq:equi}
\end{equation}
where $v\in \mathbb{R}^{3\times C}$ represents the feature output from the equivariant network for each point. 

\subsection{Local-consistent Transformation Learning}
To mitigate geometry perturbation caused by local inconsistency of previous LRF, LocoTrans proposes LCRF to perform a local-consistent transformation. The RPR module is also presented to further restore the relative pose.
We introduce these modules separately.
\label{sec:LocoTrans}

\noindent{\bf Local-consistent Reference Frame.}
Given a reference point $p_r$ in a point cloud, 
its LRF $U_r\in\mathbb{R}^{3\times 3}$ is composed of three orthonormal basis:
\begin{equation}
    \begin{aligned}
        U_r = [u_{r,1},u_{r,2}, u_{r,1} \times u_{r,2}],
    \end{aligned}
\end{equation}
where $u_{r,1}$ and $u_{r,2}$ are orthogonal normalized vectors. 
The LRF is widely used to construct local rotation-invariant representation based on each reference point and its corresponding neighbors.

\begin{figure}[htbp]
  \centering
  \includegraphics[width=\linewidth]{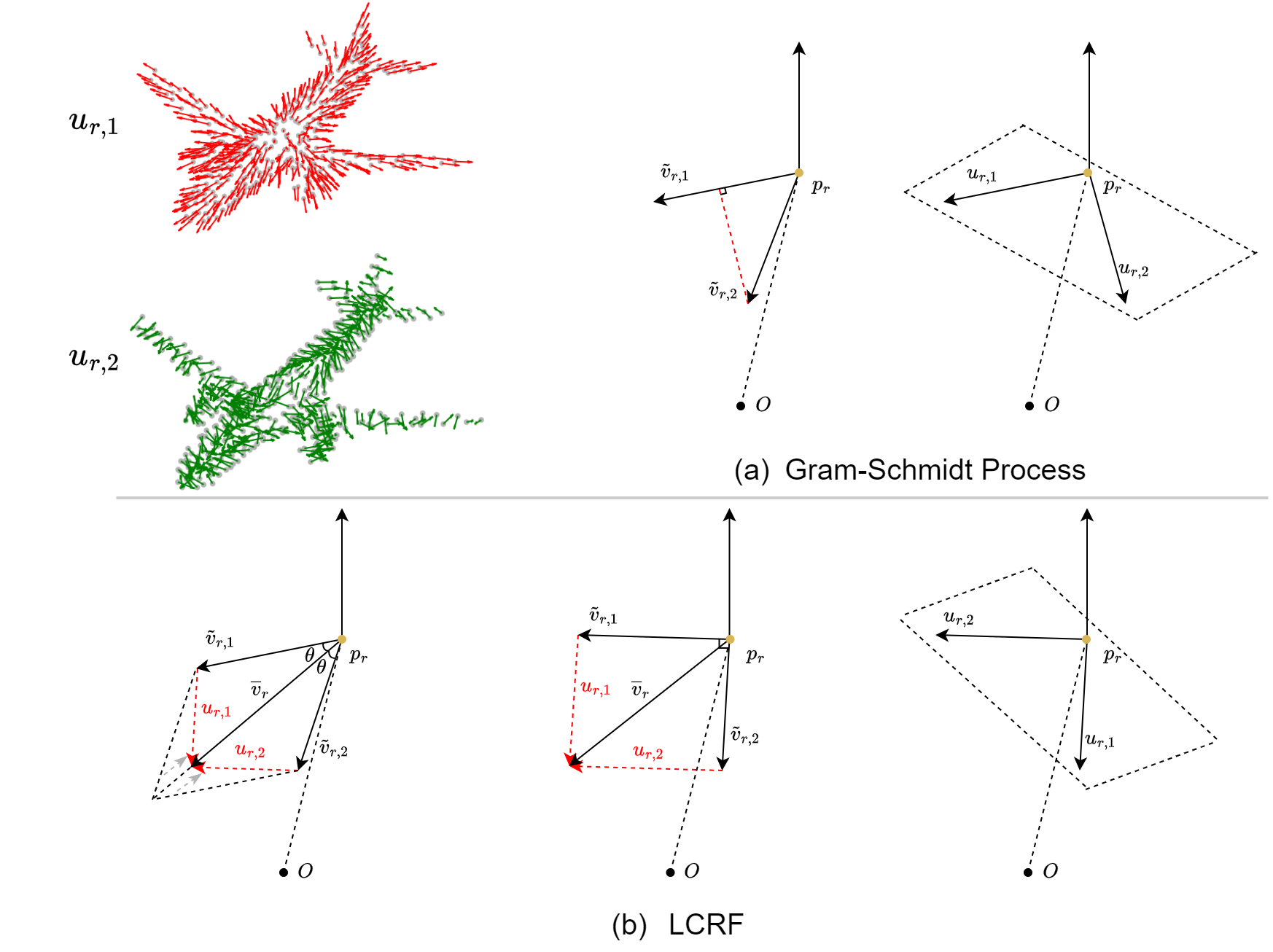} 
      \vspace{-15pt}
      \caption{Comparison between the ways of generating LRF with equivariant features. (a) Gram-Schmidt process. (b) Our LCRF. The two red dashed arrows in (b) are orthogonal. We present the visualized results of $u_{r,1}$ and $u_{r,2}$ in LRF generated by Gram-Schmidt process.}
  \label{fig:lcrf} 
  \vspace{-15pt}
\end{figure}

To leverage equivariant features for constructing a LRF, these features are first projected to $\tilde{v}_{r}=[\tilde{v}_{r,1},\tilde{v}_{r,2}]\in\mathbb{R}^{3\times 2}$ with a Multi-Layer Perceptron (MLP).
The resulting normalized vectors, $\tilde{v}_{r,1}$ and $\tilde{v}_{r,2}$, are then employed to define $u_{r,1}$ and $u_{r,2}$ within the LRF through the Gram-Schmidt process, as outlined below: 
\begin{equation}
    \begin{aligned}
        u_{r,1} = \tilde{v}_{r,1},
    \end{aligned}
\end{equation}
\begin{equation}
    \begin{aligned}
        u_{r,2} = \frac{\tilde{v}_{r,2} - \langle \tilde{v}_{r,2},u_{r,1} \rangle u_{r,1}}{\Vert \tilde{v}_{r,2} - \langle \tilde{v}_{r,2},u_{r,1} \rangle u_{r,1}\Vert},
    \end{aligned}
\end{equation}
where $\Vert\cdot\Vert$ denotes the vector norm and $\langle \cdot,\cdot \rangle$ represents the inner product. This process is depicted in Fig.~\ref{fig:lcrf}(a).

However, as shown in Fig.~\ref{fig:lcrf}(a), $u_{r,1}$ maintains orientation consistency in the local area, while $u_{r,2}$ does not. 
This phenomenon arises because $u_{r,1}$ and $u_{r,2}$, constructed using the Gram-Schmidt process, exhibit an imbalance in learning, stemming from disparities in their definitions.
Constructing local rotation-invariant representations with such inconsistent LRFs can lead to serious perturbation of local geometric relationships.

To address this issue, as shown in Fig~\ref{fig:lcrf}(b), our LCRF introduces a vector $\overline{v}_r$ that has the same direction as the angular bisector of $\tilde{v}_{r,1}$ and $\tilde{v}_{r,2}$ to construct the symmetric $u_{r,1}$ and $u_{r,2}$. Specifically, the length of $\overline{v}_r$ is first defined to be equal to $\sin{\theta}+\cos{\theta}$ to make $u_{r,1}$ and $u_{r,2}$ orthogonal, where $\theta$ is half the angle between $\tilde{v}_{r,1}$ and $\tilde{v}_{r,2}$. 
Then $u_{r,1}$ and $u_{r,2}$ are obtained by subtracting $\tilde{v}_{r,1}$ and $\tilde{v}_{r,2}$ from $\overline{v}_r$ and normalizing to the unit vector. This process can be written as: 
\begin{equation}
    \begin{aligned}
        \sin{\theta} = \sqrt{\frac{1-\tilde{v}_{r,1}^T\tilde{v}_{r,2}  }{2}},\cos{\theta} = \sqrt{\frac{1+{\tilde{v}_{r,1}}^T\tilde{v}_{r,2} }{2}},
    \end{aligned}
    \label{eq:LCRF1}
\end{equation}
\begin{equation}
    \begin{aligned}
        \overline{v}_r = \frac{(\tilde{v}_{r,1}+\tilde{v}_{r,2})}{{\Vert \tilde{v}_{r,1}+\tilde{v}_{r,2} \Vert}} (\sin{\theta}+\cos{\theta}),
    \end{aligned}
    \label{eq:LCRF2}
\end{equation}
\begin{equation}
    \begin{aligned}
        u_{r,1} = \frac{\overline{v}_r-\tilde{v}_{r,1}}{{\Vert \overline{v}_r-\tilde{v}_{r,1} \Vert}},u_{r,2} = \frac{\overline{v}_r-\tilde{v}_{r,2}}{{\Vert \overline{v}_r-\tilde{v}_{r,2} \Vert}}.
    \end{aligned}
    \label{eq:LCRF3}
\end{equation}
Analysis of the orthogonality of $u_{r,1}$ and $u_{r,2}$ can be found in the Supplementary. Although LCRF achieves the learning balance between $u_{r,1}$ and $u_{r,2}$, it still needs to be optimized for local consistency.
Given a set of reference points $P_r$ in the point cloud $P$, for two reference points $p_a, p_b \in P_r$ within the local area, we construct LRFs with normalized vectors $\tilde{v}_{a,1},\tilde{v}_{a,2}$ and $\tilde{v}_{b,1},\tilde{v}_{b,2}$ from equivariant network, respectively. The LRFs constructed by the above process for $p_a$ and $p_b$ can be represented as $U_a = [u_{a,1},u_{a,2}, u_{a,1} \times u_{a,2}]$ and $U_b = [u_{b,1},u_{b,2}, u_{b,1} \times u_{b,2}]$.
The cosine of the angle between their axes is defined as the consistency metric between $U_a$ and $U_b$. Taking $u_{a,1}$ and $u_{b,1}$ as an example, the computation is as follows:
\begin{equation}
    \begin{aligned}
        u_{a,1}^T{u_{b,1}} = \frac{(\overline{v}_a-\tilde{v}_{a,1})^{T}(\overline{v}_b-{\tilde{v}_{b,1}})}{{\Vert \overline{v}_a-\tilde{v}_{a,1} \Vert\Vert \overline{v}_b-\tilde{v}_{b,1} \Vert}}, u_{a,1}^T{u_{b,1}} \in \left( -1,1 \right),
    \end{aligned}
    \label{eq:cos_dist}
\end{equation}
where 
$\overline{v}_a$ and $\overline{v}_b$ are obtained by Eq.~\ref{eq:LCRF1} and \ Eq.~\ref{eq:LCRF2} with $\tilde{v}_{a,1},\tilde{v}_{a,2}$ and $\tilde{v}_{b,1},\tilde{v}_{b,2}$ as input.

To learn the local-consistent LRF, we design a two-step optimization strategy to maximize the consistency. 
First, to better maximize the consistency of $u_{a,1}$ and $u_{b,1}$, we simplify the representation of $ u_{a,1}^T{u_{b,1}}$. Observing the definitions of $u_{a,1}$ and $u_{b,1}$, we find that $u_{a,1}$ and $u_{b,1}$ can be defined with just one of the equivariant vectors when two equivariant vectors are orthogonal. 
Hence,
for two reference points 
$p_a$ and $p_b$, we encourage the two normalized vectors $(\tilde{v}_{1},\tilde{v}_{2})\in \{ (\tilde{v}_{a,1},\tilde{v}_{a,2}),(\tilde{v}_{b,1},\tilde{v}_{b,2}) \}$ from equivariant network to be orthogonal by an orthogonality loss:
\begin{equation}
    \begin{aligned}
        \mathcal{L}_{orth} = {\tilde{v}_{1}}^T{\tilde{v}_{2}}.
    \end{aligned}
\end{equation}

Once $\tilde{v}_{1}$ and $\tilde{v}_{2}$ are orthogonal, as shown in the center of Fig.~\ref{fig:lcrf}(b), Eq.~\ref{eq:cos_dist} can be rewritten as:
\begin{equation}
    \begin{aligned}
        u_{a,1}^T{u_{b,1}}= \tilde{v}_{a,2}^T\tilde{v}_{b,2}.
    \end{aligned}
\end{equation}
In this way, maximizing the consistency between $u_{a,1}$ and $u_{b,1}$ is converted to maximizing the consistency between $\tilde{v}_{a,2}$ and $\tilde{v}_{b,2}$. 
Next, we define consistency loss $\mathcal{L}_{consist}$ to make $\tilde{v}_{1}$ and $\tilde{v}_{2}$ learn from each other, thereby maximizing both ${\tilde{v}_{a,1}}^T{\tilde{v}_{b,1}}$ and ${\tilde{v}_{a,2}}^T{\tilde{v}_{b,2}}$ as follows:
\begin{equation}
    \begin{aligned}
        \mathcal{L}_{consist} & = ({\tilde{v}_{a,1}}^T{\tilde{v}_{b,1}}-{\tilde{v}_{a,2}}^T{\tilde{v}_{b,2}})^{2}. 
    \end{aligned}
\end{equation}
Maximizing the consistency between $u_{a,2}$ and $u_{b,2}$ can also be achieved by the above process. With our two-step optimization strategy, we can effectively enhance local consistency for both axes of LRF to alleviate the changes in relative pose.

\noindent{\bf Relative Pose Recovery.}
We can generate local rotation-invariant representation via the constructed LCRF and extract features with convolution operations. 
Taking DGCNN's edge convolution~\cite{dgcnn} as an example, two consecutive convolution operations are defined as follows:
\begin{equation}
    \begin{aligned}
        x_{r} = \mathop{\max}\limits_{j\in \mathcal{N}(p_r)}\psi(U_r^Tp_{r},U_r^T(p_{j}-p_{r})),
    \end{aligned}
    \label{eq:edgeconv0}
\end{equation}
\begin{equation}
    \begin{aligned}
        x_{r}^{'} = \mathop{\max}\limits_{j\in \mathcal{N}(x_r)}\phi(x_{r},x_{j}-x_{r}),
    \end{aligned}
    \label{eq:edgeconv}
\end{equation}
where $\mathcal{N}(\cdot)$ represents the neighbors around the input with KNN operation. $x_{r}$ represents the feature corresponding to the reference point $p_r$, and $x_{j}$ means the feature corresponding to $p_j$, the neighbor of $p_r$. Both $\phi$ and $\psi$ are MLPs. Analysis of how Eq.~\ref{eq:edgeconv0} achieves rotation invariance can be found in the Supplementary.

In Eq.~\ref{eq:edgeconv0}, each reference point and its neighbors are transformed with respect to our LCRF. However, 
as the consistency is only maintained in local regions, the relative pose between adjacent points still changes in the intermediate layers of the network.
Previous works~~\cite{zhang2020learning,Chen_2022_CVPR} have introduced a dynamic kernel $W_j$ for each neighbor in the local patch, aiming to restore the original relative pose between neighbors and the reference point:
\begin{equation}
    \begin{aligned}
        \hat{x}_{j} = W_j(\mathcal{F}_{r}^{j})x_{j},
    \end{aligned}
\end{equation}
where $\mathcal{F}_{r}^{j}$ represents the original relative pose between each neighbor and the reference point. The refined edge convolution is then given by:
\begin{equation}
    \begin{aligned}
        x_{r}^{'} = \mathop{\max}\limits_{j\in \mathcal{N}(x_r)}\phi(x_{r},\hat{x}_{j}-x_{r}).
    \end{aligned}
\end{equation}

These methods obtain $\mathcal{F}_{r}^{j}$ by hand-crafted features such as the rotation matrix and translation vector~\cite{zhang2020learning} and point pair feature~\cite{Chen_2022_CVPR}.
However, these hand-crafted features may not adequately represent complex geometric relationships for constructing the relative pose.
Hence, we introduce the RPR module, utilizing an equivariant network to capture essential geometric relationships for restoring the relative pose. 
The equivariant network can process the rotated point cloud without altering the relationship between local patches. The features extracted from it are employed to encode the correct relative pose:

\begin{equation}
    \begin{aligned}
       \mathcal{F}_{r}^{j} =U_r^{T}(v_{j}-v_{r}),
    \end{aligned}
\end{equation}
where $v_{r}$ and $v_{j}$ represent the equivariant features corresponding to $p_r$ and $p_j$, respectively. 
As shown in Eq.~\ref{eq:equi}, equivariant features preserve the rotation matrix when input data is rotated, 
necessitating the use of the LRF to eliminate rotation.

\noindent{\bf Training Losses.}
We input the features from the invariant branch, equivariant branch, and fusion layer to three different classifiers separately for classification or segmentation.
The total loss of our fusion network $\mathcal{L}$ can be written as:
\begin{equation}
    \mathcal{L}= \mathcal{L}_{i} + \mathcal{L}_{e} + \mathcal{L}_{f}+  \lambda_{orth}\mathcal{L}_{orth}+ \lambda_{consist}\mathcal{L}_{consist},
\end{equation}
where $\mathcal{L}_{i}$, $\mathcal{L}_{e}$, and $\mathcal{L}_{f}$ are the cross-entropy loss for predictions from the invariant branch, equivariant branch, and fusion layer, respectively. $\lambda_{orth}$ and $\lambda_{consist}$ are the corresponding hyperparameters to weight $\mathcal{L}_{orth}$ and $\mathcal{L}_{consist}$.
\section{Experiment}
    We evaluate the effectiveness of our method on three tasks: 3D shape classification on ModelNet40~\cite{wu20153d} dataset (Sec.~\ref{sec:3dcls}),  real-world shape classification on ScanObjectNN~\cite{Uy_2019_ICCV} dataset (Sec.~\ref{sec:rwcls}), and part segmentation on ShapeNetPart~\cite{yi2016scalable} dataset (Sec.~\ref{sec:partseg}). Moreover, we provide comprehensive ablation studies and visualization of LCRF in Sec.~\ref{sec:ablation} and Sec.~\ref{sec:visualization}. More results can be found in the Supplementary.
\subsection{Implemented details}
We use the SGD optimizer and set the initial learning rate to 0.1. 
For classification tasks, the learning rate will be adjusted with cosine annealing~\cite{loshchilov2016sgdr}, and the batch size is set to 32; for the part segmentation task, the learning rate is scaled by 0.3 every 30 epochs, and the batch size is set to 16. The maximum training epoch of all tasks is set to 250. 

To evaluate the performance of our method under various rotations, 
we follow existing training/test settings~\cite{Esteves_2018_ECCV}, i.e., z/z,
z/SO(3)\footnote{It means we apply rotation in vertical axis during training while rotating the point cloud arbitrarily for testing. The other two have similar definitions.}, and SO(3)/SO(3), to conduct experiments. Here z denotes that the input data is rotated around the vertical axis while SO(3) means rotating input arbitrarily.

\subsection{3D Shape Classification}
\label{sec:3dcls}
\noindent{\bf Dataset.}
ModelNet40~\cite{wu20153d} is widely used in the 3D shape analysis task, which provides 12,311 CAD models from 40 object categories. The dataset is split into 9,843 samples for training and 2,468 samples for testing. We follow prior work~\cite{Qi_2016_CVPR} to randomly extract 1,024 3D points from each sample for both training and testing.

\noindent{\bf Results.}
Existing methods can be categorized into three groups: rotation-sensitive, rotation-equivariant, and rotation-invariant methods. We compare our method with them in the Tab.~\ref{tab:modelnet40}. The results show that our method achieves consistent performance under three training/test settings and outperforms all methods under z/SO(3) and SO(3)/SO(3). For rotation-sensitive methods, their performance excels in z/z but drops significantly in the z/SO(3) setting. 
Although in SO(3)/SO(3) setting this issue is alleviated with additional data augmentation, a notable gap still remains in comparison with rotation-invariant methods.
This indicates that these methods struggle to handle unseen rotations during testing. Rotation-equivariant methods are robust to rotations, but their performance is restricted by their linear network structure. 
In comparison with previous rotation-invariant methods, 
our method can achieve rotation invariance while better preserving the local geometry information by constructing local-consistent reference frames and restoring the relative pose, leading to improved performance.
\vspace{-5pt}
\begin{table}[htbp] 
\begin{center}
\resizebox{\linewidth}{!}{
\begin{tabular}{c|ccccc}
\hline
\multicolumn{1}{l|}{}                 & Method            & Input & z/z  & z/SO(3) & SO(3)/SO(3) \\ \hline
\multirow{4}{*}{Rotation-sensitive}   & PointNet~\cite{Qi_2017_CVPR} (2017)   & pc     & 89.2 & 16.4    & 75.5        \\
                                      & PointNet++~\cite{NIPS2017_d8bf84be} (2017) & pc     & 89.3 & 28.6    & 85.0          \\
                                      & PointNet++~\cite{NIPS2017_d8bf84be} (2017) & pc+n   & 91.8 & 18.4    & 77.4        \\
                                      & DGCNN~\cite{dgcnn} (2019)      & pc     & 92.2 & 20.6    & 81.1        \\ \hline
\multirow{4}{*}{Rotation-equivariant} & TFN~\cite{thomas2018tensor} (2018)        & pc     & 88.5 & 85.3    & 87.6        \\
                                      & TFN-NL~\cite{Poulenard_2021_CVPR} (2021)     & pc     & 89.7 & 89.7    & 89.7        \\
                                      & VN-DGCNN~\cite{Deng_2021_ICCV} (2021)   & pc     & 89.5 & 89.5    & 90.2        \\
                                      & SVNet-DGCNN~\cite{su2022svnet} (2022)   & pc     & 90.3 & 90.3    & 90.0        \\ \hline
\multirow{10}{*}{Rotation-invariant}  & SF-CNN~\cite{Rao_2019_CVPR} (2019)     & pc     & 91.4 & 84.8    & 90.1        \\
                                      & RI-CNN~\cite{zhang-riconv-3dv19} (2019)     & pc     & 86.5 & 86.4    & 86.4        \\
                                      & RI-GCN~\cite{NEURIPS2020_5d0cb12f} (2020)     & pc     & 89.5 & 89.5    & 89.5        \\
                                      & AECNN~\cite{zhang2020learning} (2020)         & pc     & 91.0 & 91.0    & 91.0        \\
                                      & SGMNet~\cite{Xu_2021_ICCV} (2021)     & pc     & 90.0   & 90.0      & 90.0          \\
                                      & ~\citet{Li_2021_ICCV} (2021)         & pc     & 90.2 & 90.2    & 90.2        \\
                                      & PaRI-Conv~\cite{Chen_2022_CVPR} (2022)  & pc     & 91.4 & 91.4    & 91.4        \\
                                      & OrientedMP~\cite{Luo_2022_CVPR} (2022)  & pc     & 88.4 & 88.4    & 88.9        \\
                                      & PaRot~\cite{Zhang2023PaRotPR} (2023)      & pc     & 90.9 & 91.0      & 90.8        \\
                                      & Ours              & pc     & 91.6 & 91.6    & 91.5        \\ \hline
\end{tabular}
}
\vspace{-5pt}
\caption{Classification accuracy (\%) on ModelNet40 dataset. `pc'
and `n' denote the 3D coordinates and normal vectors of input data, respectively.}
\vspace{-20pt}
\label{tab:modelnet40}
\end{center}
\end{table}
\subsection{Real-world Shape Classification}
\label{sec:rwcls}
\noindent{\bf Dataset.}
ScanObjetNN~\cite{Uy_2019_ICCV} is a real-world dataset containing 15,000 incomplete objects scanned from 2,902 real-world objects. To test the effectiveness of our method in the real-world scenario, we choose the OBJ\_BG subset of ScanObjetNN which contains background
noise. The subset consists of 2,890 samples in 15 categories, which are split into 2,312 samples for training and 578 samples for testing. 

\noindent{\bf Results.}
As shown in Tab.~\ref{tab:scanobjectnn}, for the challenging real-world scenario, our method still achieves 
the best performance in all three settings. It is worth noting that PaRI-Conv~\cite{Chen_2022_CVPR}, which relies on hand-crafted LRF, exhibits poorer performance than its performance on ModelNet40. 
The effectiveness of hand-crafted LRF is limited in real-world applications for the use of geometry information such as the global center that is sensitive to background noise.
In comparison, we construct LRF with features output by equivariant branch to alleviate the impact of background noise. Although efficient, a certain absolute accuracy difference (0.5\%) occurs when the training setting is different (z/SO(3) v.s. SO(3)/SO(3)). We guess that this is because different training augmentations change the randomness of the network, which will affect learning-based LCRF.
The above results demonstrate our method can also effectively tackle real-world rotations.
\begin{table}[htbp]
\begin{center}
\resizebox{\linewidth}{!}{
\begin{tabular}{c|cccc}
\hline
Method            & Input & z/z  & z/SO(3) & SO(3)/SO(3) \\ \hline
PointNet~\cite{Qi_2017_CVPR} (2017)   & pc     & 73.3 & 16.7    & 54.7        \\
PointNet++~\cite{NIPS2017_d8bf84be} (2017) & pc     & 82.3 & 15.0    & 47.4        \\
DGCNN~\cite{dgcnn} (2019)      & pc     & 82.8 & 17.7    & 71.8        \\
\hline
RI-CNN~\cite{zhang-riconv-3dv19} (2019)     & pc     & -    & 78.4    & 78.1        \\
~\citet{Li_2021_ICCV} (2021)        & pc     & 84.3 & 84.3    & 84.3        \\
PaRI-Conv~\cite{Chen_2022_CVPR} (2022)  & pc     & 77.8 & 77.8    & 78.1        \\
PaRot~\cite{Zhang2023PaRotPR} (2023)      & pc     & -    & 82.1    & 82.6        \\
Ours              & pc     & 85.0 & 85.0    & 84.5        \\ \hline
\end{tabular}
}
\vspace{-5pt}
\caption{Classification accuracy (\%) on ScanObjectNN dataset.}
\vspace{-25pt}
\label{tab:scanobjectnn}
\end{center}
\end{table}
\subsection{Shape Part Segmentation}
\label{sec:partseg}
\begin{figure}[htbp]
  \centering
  \includegraphics[width=\linewidth]{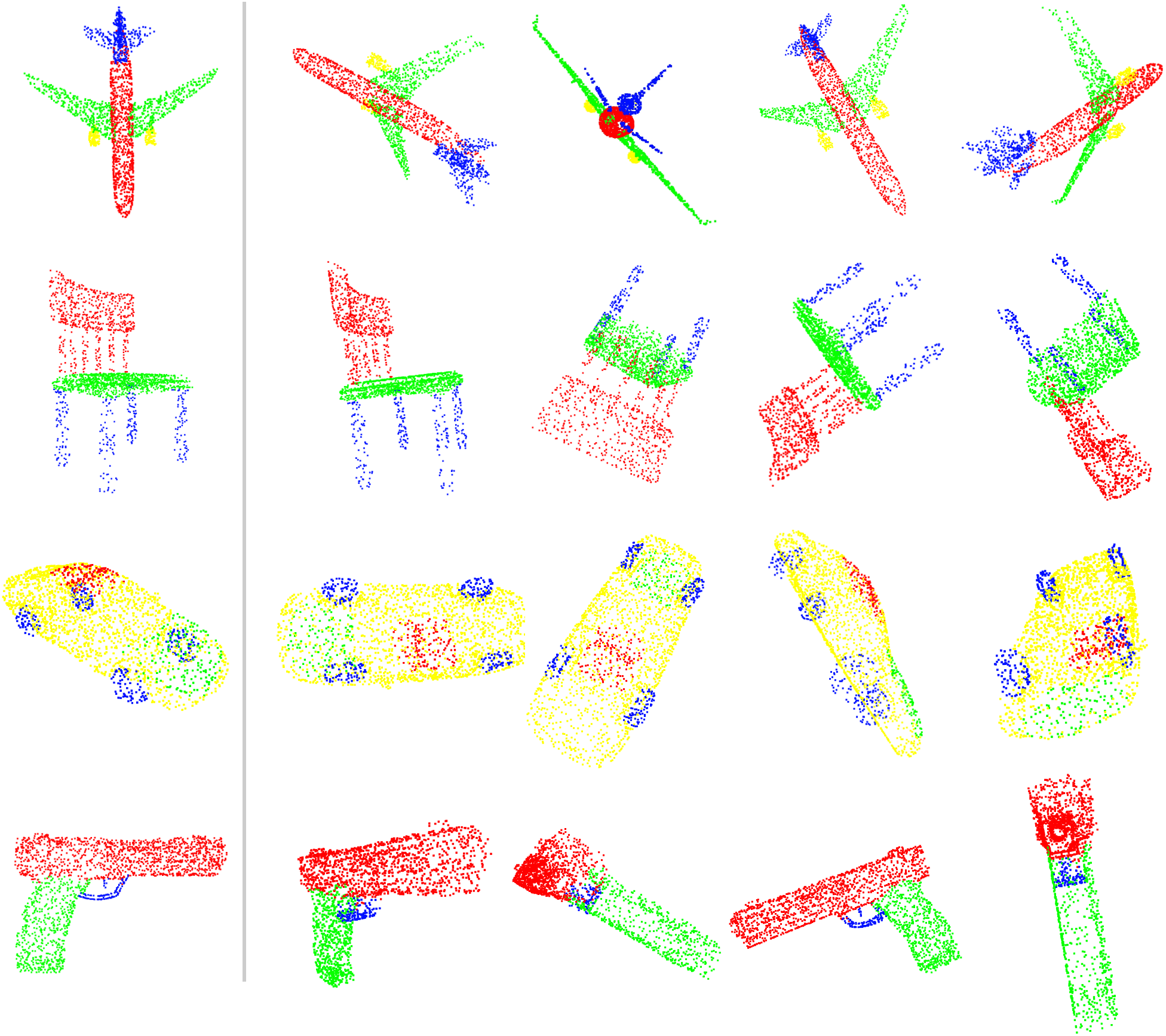}
      \caption{Visualization of part segmentation results on ShapeNetPart dataset under z/SO(3) setting. From top to bottom, we present the segmentation results for four categories: airplane, chair, car, and pistol. The leftmost column is the ground truth and the right four columns are the predictions of our network under arbitrary rotations. 
      } 
  \label{fig:partseg} 
  \vspace{-10pt}
\end{figure}

\noindent{\bf Dataset.}
We adopt ShapeNetPart dataset~\cite{yi2016scalable} to evaluate our method on the shape part segmentation task. ShapeNetPart contains 16, 881 3D samples from 16 categories, which are further subdivided into 50 parts. 
We refer to the data split in ~\cite{Qi_2017_CVPR} and sample 2,048 points for each 3D object.
\noindent{\bf Results.}
In order to fairly compare with existing methods, we report our results with mean Intersection-over-Union (mIoU) over all instances and all classes together in Tab.~\ref{tab:shapenet}. The results show our method is competitive on Class mIOU and achieves the best performance on Insta. mIOU. Furthermore, compared to our backbone DGCNN, LocoTrans significantly boosts the performance by 42.7\% in z/SO(3) setting.
We also give visualized segmentation results of 3D objects in Fig.~\ref{fig:partseg}, and the figure indicates our network is robust to arbitrary rotations in segmentation.
\vspace{-5pt}
\begin{table}[htbp]
\begin{center}
\resizebox{\linewidth}{!}{
\begin{tabular}{c|ccccc}
\cline{1-6}
\multirow{2}{*}{Method} & \multicolumn{1}{l}{\multirow{2}{*}{Input}} & \multicolumn{2}{c}{z/SO(3)}                                      & \multicolumn{2}{c}{SO(3)/SO(3)}                                  \\ \cline{3-6} 
                        & \multicolumn{1}{l}{}                        & \multicolumn{1}{l}{Class mIOU} & \multicolumn{1}{l}{Insta. mIOU} & \multicolumn{1}{l}{Class mIOU} & \multicolumn{1}{l}{Insta. mIOU} \\ \hline
PointNet~\cite{Qi_2017_CVPR} (2017)         & pc                                          & 37.8                           & -                               & 74.4                           & -                               \\
PointNet++~\cite{NIPS2017_d8bf84be} (2017)       & pc                                          & 48.2                           & -                               & 76.7                           & -                               \\
DGCNN~\cite{dgcnn} (2019)           & pc                                          & 37.4                           & -                               & 73.3                           & -                               \\ \hline
RI-CNN~\cite{zhang-riconv-3dv19} (2019)           & pc                                          & 75.3                           & -                               & 75.5                           & -                               \\
AECNN~\cite{zhang2020learning} (2020)            & pc                                          & 80.2                           & -                               & 80.2                           & -                               \\
RI-GCN~\cite{NEURIPS2020_5d0cb12f} (2020)           & pc                                          & 77.2                           & -                               & 77.3                           &-                                 \\
VN-DGCNN~\cite{Deng_2021_ICCV} (2021)           & pc                                          & -                           & 81.4                               & -                           &  81.4                               \\
~\citet{Li_2021_ICCV} (2021)               & pc                                          & 74.1                           & 81.7                            & 74.1                           & 81.7                            \\
PaRI-Conv~\cite{Chen_2022_CVPR} (2022)        & pc                                          & -                              & 83.8                            & -                              & 83.8                            \\
PaRot~\cite{Zhang2023PaRotPR} (2023)            & pc                                          & 79.2                           & -                               & 79.5                           & -                               \\
Ours                    & pc                                          & 80.1                           & 84.0 
& 80.0                               & 83.8 \\       \hline
\end{tabular}
}
\vspace{-5pt}
\caption{Part segmentation results on ShapeNetPart dataset. ‘Class mIOU’ stands for mIoU (\%) over all classes while ‘Insta. mIOU’ denotes mIoU over all instances.}
\vspace{-30pt}
\label{tab:shapenet}
\end{center}
\end{table}
\subsection{Ablation Studies}
\label{sec:ablation}
In this section, we conduct experiments to demonstrate the effectiveness of each component in LocoTrans. Further ablation studies are also designed on the LCRF and the RPR module for deep analysis. If no otherwise specified, all experiments are performed on ModelNet40 dataset under the z/SO(3) setting and reported with accuracy (\%).

\vspace{2mm}
\noindent{\bf Effects of Different Components.}
We carry out a series of experiments to validate our method and report the performance in Tab.~\ref{tab:ablation}. 
We take the output of the invariant branch of our network as the Baseline in Row \#1 without using LCRF and RPR module. Aggregation in Row \#2 refers to fusing the output of the invariant branch and the equivariant branch, which contributes to the improvement of classification accuracy. 
In Row \#3 we replace the LRF built through the Gram-Schmidt process with our LCRF. Compared to Row \#2, our LCRF improves the performance by alleviating the perturbation of local geometry with local-consistent transformation. 
We also test the performance of the RPR module with the original coordinates and the equivariant features as input.
The results in Row \#4 and Row \#5 show both variations can enhance the performance of the network for addressing the loss of relative pose. 
Notably, the comparison between Row \#4 and Row \#5 highlights that features extracted by the equivariant branch outperform original coordinates in representing relative pose. 
Finally, our network performs optimally when all modules are in use.
\begin{table}[tbp]
\begin{center}
\resizebox{\linewidth}{!}{
\begin{tabular}{c|ccccc|c}
\hline
\multirow{2}{*}{Row} & \multirow{2}{*}{Baseline} & \multirow{2}{*}{Aggregation} & \multirow{2}{*}{LCRF} & \multicolumn{2}{c|}{RPR} & \multirow{2}{*}{z/SO(3)} \\ \cline{5-6}
                     &                                    &                                 &                       & coordinate     & feature    &                          \\ \hline
\#1                    & \checkmark                                  &                                 &                       &                &            & 89.1                    \\
\#2                    &                                    & \checkmark                              &                       &                &            & 90.4                     \\
\#3                    &                                    & \checkmark                               & \checkmark                     &                &            & 90.7                     \\
\#4                    &                                    & \checkmark                               &                       & \checkmark              &            & 90.8                     \\
\#5                    &                                    & \checkmark                               &                       &                & \checkmark          & 91.3                     \\
\#6                    &                                    & \checkmark                               & \checkmark                     &                & \checkmark          & 91.6                     \\ \hline
\end{tabular}
}
\caption{Ablation study of our method on ModelNet40 dataset in z/SO(3) setting.‘coordinate’ denotes the coordinates of point cloud while ‘feature’ represents the equivariant features.}
\label{tab:ablation}
\vspace{-20pt}
\end{center}
\end{table}

\vspace{2mm}
\noindent{\bf Effects of LCRF.}
We also conduct experiments to explore the role of different LRFs in local rotation-invariant representation construction in Tab.~\ref{tab:LCRF}. 
In Row \#1, we handcraft LRF with the global center and local barycenter, while leveraging features from the equivariant network to construct LRF based on the Gram-Schmidt process in Row \#2. 
Due to the lack of local consistency, generating local rotation-invariant representation with either hand-crafted LRF or learning-based LRF built by Gram-Schmidt process results in worse performance. Our LCRF brings improvement by achieving the learning balance between two vectors from an equivariant network. Moreover, comparing the results on ModelNet40 and ScanObjectNN, LCRF brings more significant improvement in the latter containing background noise because data in the former is clean and easier to classify.
\vspace{-5pt}
\begin{table}[htbp]
\begin{center}
\footnotesize
\begin{tabular}{c|ccc}
\hline
\multirow{2}{*}{Row} & \multirow{2}{*}{LRF} & \multicolumn{2}{c}{z/SO(3)} \\ \cline{3-4} 
                     &                      & ModelNet40  & ScanObjectNN  \\ \hline
\#1                    & hand-crafted         & 91.2        & 82.8          \\
\#2                    & Gram-Schmidt process & 91.3        & 84.2          \\
\#3                    & LCRF                 & 91.6        & 85.0          \\ \hline
\end{tabular}
\vspace{-5pt}
\caption{Ablation study on LRF construction.}
\vspace{-15pt}
\label{tab:LCRF}
\end{center}
\end{table}

\begin{figure*}[htbp]
  \centering
  \includegraphics[width=0.85\linewidth]{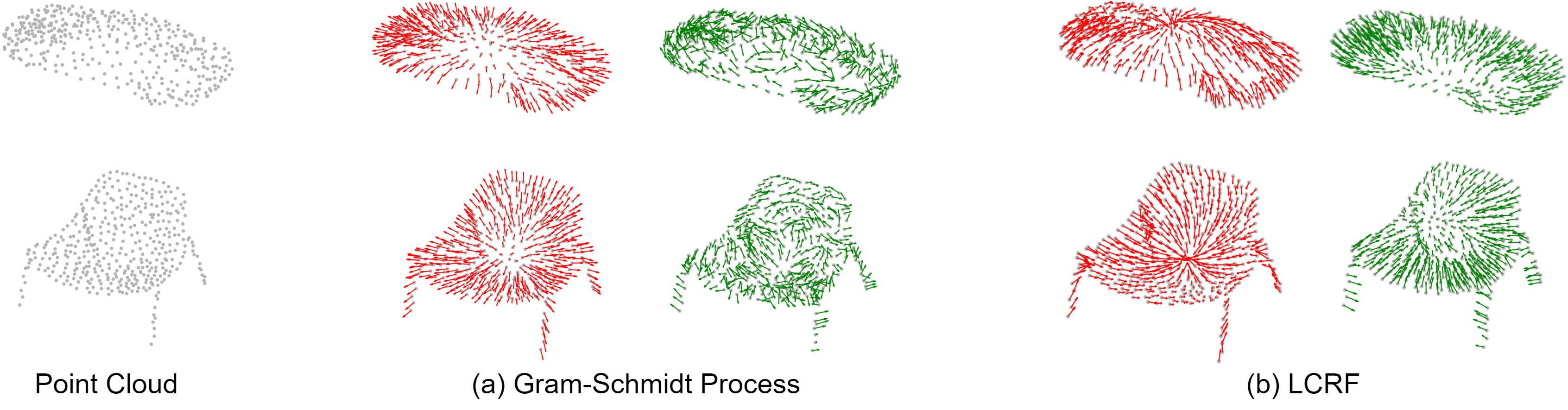} 
      \vspace{-6pt}
      \caption{Visualization of $u_r^1$ (Red) and $u_r^2$ (Green) in LRF. (a) LRF generated with the Gram-Schmidt process. (b) LCRF. While the Gram-Schmidt process fails to maintain local consistency in $u_r^2$, Our LCRF learns local-consistent orientation in both axes.
      } 
      \vspace{-15pt}
  \label{fig:visualization} 
\end{figure*}

\noindent{\bf Effects of RPR Module.}
We compare the effects of using different features to represent the relative pose in Tab.~\ref{tab:rpr}. In Row \#1 we directly utilize the coordinates of the point cloud to represent the relative pose while in Row \#2, Row \#3, and Row \#4, we use features extracted from the coordinates in different ways. In Row \#2 hand-crafted feature means the APPF-w/oDirection proposed by PaRI-Conv~\cite{Chen_2022_CVPR} which encodes the relative Euclidean distance and angle in the local patch. 
Equivariant feature in Row \#3 represents the feature extracted from the equivariant branch while invariant feature in Row \#4 represents the feature extracted from the invariant branch. 
The comparison of Row \#1 and Row \#2 shows hand-crafted features cannot improve performance because they may not adequately represent essential geometric relationships. 
In comparison, equivariant branch can better capture the local geometry information to encode relative pose, leading to the improvement of classification accuracy in Row \#3. The performance in Row \#4 declines because the invariant feature itself suffers from the relative pose issue. 
\vspace{-7pt}
\begin{table}[htbp]
\begin{center}
\begin{tabular}{c|c|c}
\hline
Row & Relative pose representation & z/SO(3) \\ \hline
\#1   & coordinate     & 91.0    \\
\#2   & hand-crafted feature      & 91.0    \\
\#3   & equivariant feature       & 91.6  \\
\#4   & invariant feature         & 90.8  \\ \hline
\end{tabular}
\vspace{-5pt}
\caption{Ablation study on relative pose representation.}
\label{tab:rpr}
\end{center}
\end{table}
\vspace{-21pt}

\noindent{\bf Computational Burden.}
We provide the number of parameters, FLOPs, and accuracy of the DGCNN-based networks in Tab.~\ref{tab:complexity}. 
The two branches of both `Ours-base' (no LCRF and no RPR, Row \#2 in Tab.~\ref{tab:ablation}) and `Ours' are built upon DGCNN (1.81M Params) and VN-DGCNN (2.89M Params).
Tab.~\ref{tab:complexity} shows that the two-branch structure indeed requires more costs. Note that, the fusion layer and projection layer in `Ours-base' and `Ours' cause extra parameters than the sum of DGCNN and VN-DGCNN. However, `Ours-base' v.s. ~\citet{Li_2021_ICCV} and `Ours-base' v.s. PaRI-Conv~\cite{Chen_2022_CVPR} reveal that naively increasing parameters does not necessarily enhance performance (even degradation occurs). The `Ours-base' v.s. `Ours' shows that while our proposed LCRF and RPR introduce minor additional costs, they significantly improve performance, highlighting the effectiveness of our proposed strategies. Nonetheless, the added equivariant branch does increase overall costs, prompting us to seek more cost-effective ways to leverage equivariant information in future research.
\vspace{-8pt}
\begin{table}[h]
\begin{center}

\scalebox{0.7}{
\begin{tabular}{c|cccc}
\hline
         &  ~\citet{Li_2021_ICCV}   & PaRI-Conv~\cite{Chen_2022_CVPR} &  Ours-base & Ours  \\ \hline
Params   &  2.91M & 1.85M    & 6.22M         & 6.27M\\
FLOPs    &  3747M & 1938M    & 7552M         & 7998M \\
Acc.(ModelNet40) & 90.2  & 91.4     & 90.4          & 91.6 \\
Acc.(ScanObjectNN) &  84.3  & 77.8     & 82.8          & 85.0 \\\hline
\end{tabular}
}
\end{center}
\vspace{-15pt}
\caption{Computational burden on the classification network.}
\vspace{-12pt}
\label{tab:complexity}
\end{table}

\subsection{Visualization of LCRF}
\label{sec:visualization}
We visualize two types of LRFs in Fig.~\ref{fig:visualization}. For the Gram-Schmidt process, it can always maintain local-consistent orientation in $u_r^1$ but not in $u_r^2$. With these kinds of LRFs, points on similar local structures are transformed into different reference frames, which is not conducive for network to learn the local structure.
In contrast, 
our LCRF maintains consistency across different axes and preserves the local geometry.

\section{Conclusion}
    In this paper, we proposed Local-consistent Transformation (LocoTrans) learning to effectively achieve local rotation-invariant representation. LocoTrans is built upon the equivariant network and consists of two modules. Specifically, the LCRF constructs local-consistent reference frames to preserve local geometry relationships when performing transformation via LRF. However, the relative pose between adjacent points still changes. To further restore relative pose, our RPR module leverages the equivariant network to encode pose information from the original local coordinates and fuses pose information with neighbor features. Experimental results demonstrate the effectiveness of our method.

\section{Acknowledgment}
    This work was supported by the National Natural Science Foundation of China under Grant 62076101, Guangdong Basic and Applied Basic Research Foundation under Grant 2023A1515010007, Guangdong Provincial Key Laboratory of Human Digital Twin under Grant 2022B1212010004, and CAAI-Huawei MindSpore Open Fund.

{
    \small
    \bibliographystyle{ieeenat_fullname}
    \bibliography{main}
}

\appendix
\clearpage
\setcounter{page}{1}
\maketitlesupplementary
In this Supplementary Material, we first provide the proof of
the properties of LCRF in Sec.~\ref{secsm:LCRF}. Additional experiments are presented in Sec.~\ref{secsm:exp} for further analysis. Next, we summarize the limitation of our work in  Sec.~\ref{secsm:limit}.
Finally, we show more visualization results on LCRF and LRF built by the Gram-Schmidt process for comparison in Sec.~\ref{secsm:vis}.

\begin{figure}[htbp]
  \centering
  \includegraphics[width=0.4\linewidth]{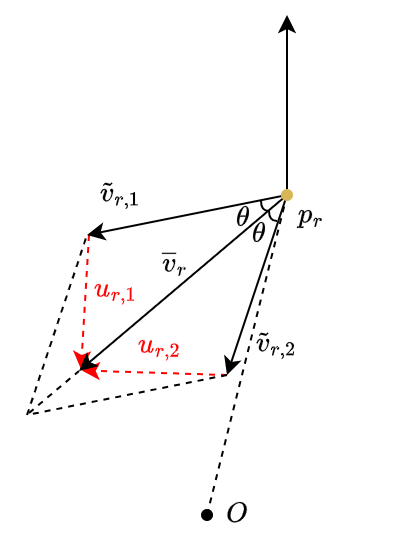} 
      \caption{Illustration of LCRF.
      } 
  \label{figsm:lcrf} 
  \vspace{-6mm}
\end{figure}
\section{Proofs of LCRF Properties}
\label{secsm:LCRF}
In this section, we analyze the orthogonality of $u_{r,1}$ and $u_{r,2}$ in LCRF and 
the rotation invariance in Eq.~\ref{eq:edgeconv0} of the Submission.
\subsection{Orthogonality of $u_{r,1}$ and $u_{r,2}$ in LCRF}
\noindent{\bf Theorem 1.} Given two normalized vectors $\tilde{v}_{r,1},\tilde{v}_{r,2}\in\mathbb{R}^{3\times 1}$, with $\theta\in(0,\frac{\pi}{2})$ being half the angle between them, $u_{r,1}$ and $u_{r,2}$ in LCRF, defined as follows, are orthogonal:
\begin{equation}
    \begin{aligned}
        \sin{\theta} = \sqrt{\frac{1-\tilde{v}_{r,1}^T\tilde{v}_{r,2}  }{2}},\cos{\theta} = \sqrt{\frac{1+{\tilde{v}_{r,1}}^T\tilde{v}_{r,2} }{2}},
    \end{aligned}
    \label{eqsm:LCRF1}
\end{equation}
\begin{equation}
    \begin{aligned}
        \overline{v}_r = \frac{(\tilde{v}_{r,1}+\tilde{v}_{r,2})}{{\Vert \tilde{v}_{r,1}+\tilde{v}_{r,2} \Vert}} (\sin{\theta}+\cos{\theta}),
    \end{aligned}
    \label{eqsm:LCRF2}
\end{equation}
\begin{equation}
    \begin{aligned}
        u_{r,1} = \frac{\overline{v}_r-\tilde{v}_{r,1}}{{\Vert \overline{v}_r-\tilde{v}_{r,1} \Vert}},u_{r,2} = \frac{\overline{v}_r-\tilde{v}_{r,2}}{{\Vert \overline{v}_r-\tilde{v}_{r,2} \Vert}}.
    \end{aligned}
    \label{eqsm:LCRF3}
\end{equation}
\textit{Proof.} 
Fig.~\ref{figsm:lcrf} illustrates our LCRF. 
The orthogonality of non-zero vectors $u_{r,1}$ and $u_{r,2}$ depends on whether $u_{r,1}^Tu_{r,2}=0$.
To demonstrate this, we first compute $u_{r,1}^Tu_{r,2}$:
\begin{equation}
    \begin{aligned}
        u_{r,1}^Tu_{r,2} = \frac{(\overline{v}_r-\tilde{v}_{r,1})^T}{{\Vert \overline{v}_r-\tilde{v}_{r,1} \Vert}}\frac{(\overline{v}_r-\tilde{v}_{r,2})}{{\Vert \overline{v}_r-\tilde{v}_{r,2} \Vert}}.
    \end{aligned}
    \label{eqsm:22}
\end{equation}
For brevity, we only focus on the numerator in Eq.~\ref{eqsm:22}, which can be written as:
\begin{equation}
    \small
    \begin{aligned}
       &(\overline{v}_r-\tilde{v}_{r,1})^T(\overline{v}_r-\tilde{v}_{r,2}) = \overline{v}_r^T\overline{v}_r - \overline{v}_r^T\tilde{v}_{r,2} -\tilde{v}_{r,1}^T\overline{v}_r \\
       &+\tilde{v}_{r,1}^T{\tilde{v}_{r,2}}.
    \end{aligned}
    \label{eqsm:23}
\end{equation}
From Eq.~\ref{eqsm:LCRF1}, $\tilde{v}_{r,1}^T{\tilde{v}_{r,2}}$ can be derived by:
\begin{equation}
    \begin{aligned}
       \tilde{v}_{r,1}^T{\tilde{v}_{r,2}} = 2{\cos}^2\theta-1.
    \end{aligned}
    \label{eqsm:24}
\end{equation}
From Eq.~\ref{eqsm:LCRF2} and Eq.~\ref{eqsm:24}, we can derive $\overline{v}_r^T\overline{v}_r$, $\overline{v}_r^T\tilde{v}_{r,2}$ and $\tilde{v}_{r,1}^T\overline{v}_r$ separately:
\begin{equation}
    \small
    \begin{aligned}
       \overline{v}_r^T\overline{v}_r &=\frac{(\tilde{v}_{r,1}+\tilde{v}_{r,2})^T(\tilde{v}_{r,1}+\tilde{v}_{r,2})}{{{\Vert \tilde{v}_{r,1}+\tilde{v}_{r,2} \Vert}^2}} (\sin{\theta}+\cos{\theta})^2 \\
       &=(\sin{\theta}+\cos{\theta})^2, \\
       \overline{v}_r^T\tilde{v}_{r,2}        &=\frac{(\tilde{v}_{r,1}+\tilde{v}_{r,2})^T{\tilde{v}_{r,2}}}{{\Vert \tilde{v}_{r,1}+\tilde{v}_{r,2} \Vert}} (\sin{\theta}+\cos{\theta}) \\
       &=\frac{(\tilde{v}_{r,1}^T{\tilde{v}_{r,2}}+1)}{\sqrt{2+2\tilde{v}_{r,1}^T{\tilde{v}_{r,2}}}}(\sin{\theta}+\cos{\theta})\\
       &=\frac{(2{\cos}^2\theta-1+1)}{\sqrt{2+2(2{\cos}^2\theta-1)}}(\sin{\theta}+\cos{\theta})\\
       &=\cos{\theta}(\sin{\theta}+\cos{\theta}), \\
       \tilde{v}_{r,1}^T\overline{v}_r
       &=\frac{{\tilde{v}_{r,1}}^T(\tilde{v}_{r,1}+\tilde{v}_{r,2})}{{\Vert \tilde{v}_{r,1}+\tilde{v}_{r,2} \Vert}} (\sin{\theta}+\cos{\theta}) \\
       &=\frac{(1+\tilde{v}_{r,1}^T{\tilde{v}_{r,2}})}{\sqrt{2+2\tilde{v}_{r,1}^T{\tilde{v}_{r,2}}}}(\sin{\theta}+\cos{\theta})\\
       &=\frac{(1+2{\cos}^2\theta-1)}{\sqrt{2+2(2{\cos}^2\theta-1)}}(\sin{\theta}+\cos{\theta})\\
       &=\cos{\theta}(\sin{\theta}+\cos{\theta}). \\
    \end{aligned}
    \label{eqsm:25}
\end{equation}
Thus, we can proceed to derive:
    \begin{equation}
    \footnotesize
    \begin{aligned}
       (\overline{v}_r-\tilde{v}_{r,1})^T(\overline{v}_r-\tilde{v}_{r,2}) 
       &= \overline{v}_r^T\overline{v}_r - \overline{v}_r^T\tilde{v}_{r,2} -\tilde{v}_{r,1}^T\overline{v}_r+\tilde{v}_{r,1}^T{\tilde{v}_{r,2}} \\
       &=(\sin{\theta}+\cos{\theta})^2-\cos{\theta}(\sin{\theta}+\cos{\theta})\\
       &-\cos{\theta}(\sin{\theta}+\cos{\theta}) + 2{\cos}^2\theta-1\\
       &=1+2\sin{\theta}\cos{\theta}-2{\cos}^2\theta-2\sin{\theta}\cos{\theta} \\
       &+2{\cos}^2\theta-1 \\
       &=0,
    \end{aligned}
    \label{eqsm:26}
\end{equation}
which can be used to derive:
\begin{equation}
    \begin{aligned}
        u_{r,1}^Tu_{r,2} = \frac{(\overline{v}_r-\tilde{v}_{r,1})^T}{{\Vert \overline{v}_r-\tilde{v}_{r,1} \Vert}}\frac{(\overline{v}_r-\tilde{v}_{r,2})}{{\Vert \overline{v}_r-\tilde{v}_{r,2} \Vert}}=0.
    \end{aligned}
\end{equation}
Therefore, $u_{r,1}$ and $u_{r,2}$ are orthogonal.
\subsection{Rotation Invariance in Eq.~\ref{eq:edgeconv0} of the Submission}
To prove the rotation invariance in Eq.~\ref{eq:edgeconv0} of the Submission, we begin with analyzing how LCRF introduces rotation invariance since the local rotation-invariant representation $U_r^{T}p$ is the input of the equation. The detailed proof is shown as follows.

\noindent{\bf Theorem 2.} 
For any reference point $p_r$ and its $K$ neighbors $\{p_j\}_{j=1}^{K}$, 
the equation defined as follows can achieve rotation invariance under arbitrary rotation $R\in$SO(3):
\begin{equation}
    \begin{aligned}
        x_{r} = \mathop{\max}\limits_{j\in \mathcal{N}(p_r)}\psi(U_r^Tp_{r},U_r^T(p_{j}-p_{r})),
    \end{aligned}
    \label{eq:edgeconv0_}
\end{equation}
where $U_r$ is LCRF, $\mathcal{N}(\cdot)$ denotes the KNN operation invariant to rotation, and $\psi$ represents a MLP. \\
\\
\textit{Proof.} 
For a point $p\in P$, when applying a random rotation matrix R to it, for an equivariant feature $v$, we have:
\begin{equation}
    \begin{aligned}
        v^{*} = h(Rp) = Rh(p)= Rv,
    \end{aligned}
\end{equation}
where $h(\cdot)$ is the equivariant network and the superscript $*$ indicates that the result corresponds to the rotated input. $\tilde{v}_{r}=[\tilde{v}_{r,1},\tilde{v}_{r,2}]$ is the equivariant feature used to construct LCRF, so it also satisfies:
\begin{equation}
    \begin{aligned}
        \tilde{v}_{r}^{*} = R \tilde{v}_{r}=[R\tilde{v}_{r,1},R\tilde{v}_{r,2}].
    \end{aligned}
\end{equation}
Therefore, for Eq.~\ref{eqsm:LCRF1}, Eq.~\ref{eqsm:LCRF2} and Eq.~\ref{eqsm:LCRF3} we have:
\begin{equation}
    \small
    \begin{aligned}
        &{\sin}^{*}\theta = \sqrt{\frac{1-({R\tilde{v}_{r,1})}^TR\tilde{v}_{r,2}}{2}}= \sqrt{\frac{1-{\tilde{v}_{r,1}}^T\tilde{v}_{r,2}}{2}}=\sin{\theta},\\
        &{\cos}^{*}\theta = \sqrt{\frac{1+({R\tilde{v}_{r,1})}^TR\tilde{v}_{r,2}}{2}}= \sqrt{\frac{1+{\tilde{v}_{r,1}}^T\tilde{v}_{r,2}}{2}}=\cos{\theta},
    \end{aligned}
\end{equation}
\begin{equation}
    \begin{aligned}
        \overline{v}_r^{*} 
        &= \frac{(R\tilde{v}_{r,1}+R\tilde{v}_{r,2})}{{\Vert R\tilde{v}_{r,1}+R\tilde{v}_{r,2} \Vert}} ({\sin}^{*}\theta+{\cos}^{*}\theta) \\
        &= \frac{R(\tilde{v}_{r,1}+\tilde{v}_{r,2})}{{\Vert \tilde{v}_{r,1}+\tilde{v}_{r,2} \Vert}} (\sin{\theta}+\cos{\theta}) \\
        &=R\overline{v}_r,
    \end{aligned}
\end{equation}
\begin{equation}
    \begin{aligned}
        &u_{r,1}^{*} = \frac{\overline{v}_r^{*}-R\tilde{v}_{r,1}}{{\Vert \overline{v}_r^{*}-R\tilde{v}_{r,1} \Vert}}=\frac{R(\overline{v}_r-\tilde{v}_{r,1})}{{\Vert \overline{v}_r-\tilde{v}_{r,1} \Vert}}=Ru_{r,1},  \\
        &u_{r,2}^{*} = \frac{\overline{v}_r^{*}-R\tilde{v}_{r,2}}{{\Vert \overline{v}_r^{*}-R\tilde{v}_{r,2} \Vert}} = \frac{R(\overline{v}_r-\tilde{v}_{r,2})}{{\Vert \overline{v}_r-\tilde{v}_{r,2} \Vert}} = Ru_{r,2}.
    \end{aligned}
\end{equation}
For LCRF $U_{r}=[u_{r,1},u_{r,2},u_{r,1}\times u_{r,2}]$, it satisfies:
\begin{equation}
    \begin{aligned}
        U_{r}^{*} 
        &=[u_{r,1}^{*},u_{r,2}^{*},u_{r,1}^{*}\times u_{r,2}^{*}] \\
        &=[Ru_{r,1},Ru_{r,2},Ru_{r,1}\times Ru_{r,2}] \\
        &= RU_{r},
    \end{aligned}
\end{equation}
which can be used to achieve rotation invariance through:
\begin{equation}
    \begin{aligned}
        {U_{r}^{*}}^TRp = {(RU_{r})}^TRp ={U_{r}}^TR^{T}Rp = U_{r}^{T}p.
    \end{aligned}
\end{equation}
Hence, we can derive: 
\begin{equation}
    \begin{aligned}
        x_{r}^{*} 
        &= \mathop{\max}\limits_{j\in \mathcal{N}(Rp_r)}\psi({U_r^{*}}^TRp_{r},{U_r^{*}}^T(Rp_{j}-Rp_{r}))\\
        &= \mathop{\max}\limits_{j\in \mathcal{N}(p_r)}\psi({U_r}^Tp_{r},{U_r}^T(p_{j}-p_{r}))\\
        &=x_{r},
    \end{aligned}
    \label{eqsm:edgeconv0}
\end{equation}
which proves Eq.~\ref{eq:edgeconv0_} can achieve rotation invariance.
\section{Additional Experiments}
\label{secsm:exp}
\subsection{Semantic Segmentation}
\noindent{\bf Dataset.}
The S3DIS~\cite{Armeni_2016_CVPR} dataset consists of 6 large-scale indoor areas with 271 rooms in 13 categories. Following previous work~\cite{Thomas_2019_ICCV}, we select Area-5 for testing, while the other five areas are used for training. 

\noindent{\bf Results.}
Given that few works perform experiments on S3DIS, we compare our method with our backbone. As shown in Tab.~\ref{tab:s3dis}, DGCNN~\cite{dgcnn} struggles to process complex scene-level data under arbitrary rotations, and our method outperforms it by a large margin, especially in the z/SO(3) setting. 
The results indicate that our LocoTrans can also work on large-scale point cloud data.
\begin{table}[htbp]
\begin{center}
\resizebox{1.0\linewidth}{!}{
\begin{tabular}{c|ccccc}
\hline
\multirow{2}{*}{Method} & \multirow{2}{*}{Input} & \multicolumn{2}{c}{z/SO(3)} & \multicolumn{2}{c}{SO(3)/SO(3)} \\ \cline{3-6} 
                        &                         & mIOU         & Accuracy          & mIOU           & Accuracy          \\ \hline
DGCNN~\cite{dgcnn}                   & pc                      & 3.0          & 18.1         & 42.8           & 80.7           \\
Ours                    & pc                      & 54.2         & 84.8         & 56.0           & 84.7           \\ \hline
\end{tabular}}
\caption{Semantic segmentation results on S3DIS dataset, which are reported by mIOU (\%) and accuracy (\%) separately.}
\label{tab:s3dis}
\end{center}
\end{table}

\subsection{Performance under Various Perturbations}
\begin{figure}[t]
  \centering
  \includegraphics[width=\linewidth]{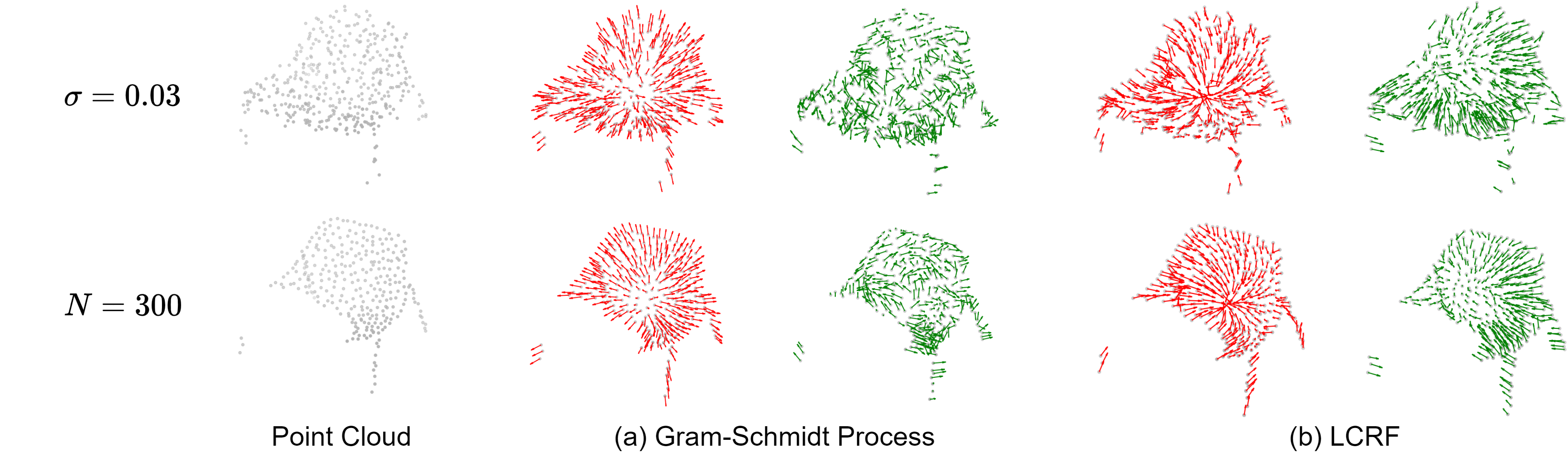}
      \caption{Visualization of $u_r^1$ (Red) and $u_r^2$ (Green) in LRF under perturbation. 
      }
  \label{fig:ptb} 
\end{figure}

In Tab.~\ref{tab:ptb},
we evaluate the performance of our LCRF and LRF built from Gram-Schmidt process under perturbations
by introducing noise and random dropout. 
Despite both methods showing decreased performance under perturbations, our LCRF consistently outperforms Gram-Schmidt process.  Moreover,  
the visualization in Fig.~\ref{fig:ptb} shows that LCRF still maintains a degree of local consistency in both $u_r^1$ (Red) and $u_r^2$ (Green). We further try to introduce noises during training to improve the robustness on noised data, getting $88.7$\% with our LCRF and $88.3$\% with LRF from Gram-Schmidt process for $\sigma=0.03$. It shows training with noises benefits the performance on noised data and our LCRF still performs better.

\begin{table}[t]
\begin{center}
\resizebox{\linewidth}{!}{
\hfill
\begin{minipage}[t]{0.5\linewidth}
\resizebox{\linewidth}{!}{
\begin{tabular}{c|ccccc}
\hline
$\sigma$                    & 0    & 0.01 & 0.02 & 0.03  \\ \hline
G-S & 91.3 & 85.9 & 74.0 & 46.0 \\
LCRF                 & 91.6 & 86.8 & 77.1 & 48.5  \\ \hline
\end{tabular}
}
\end{minipage}

\begin{minipage}[t]{0.5\linewidth}
\resizebox{\linewidth}{!}{
\begin{tabular}{c|ccccc}
\hline
$N$ & 0    & 100  & 200  & 300   \\ \hline
G-S     & 91.3 & 87.1 & 80.1 & 71.6 \\
LCRF                     & 91.6 & 87.2 & 80.3 & 72.4\\ \hline
\end{tabular}
}
\end{minipage}

\hfill
}
\caption{Classification accuracy (\%) under different scales of noises $\sigma$ (Left) and number of dropout points $N$ (Right) on ModelNet40 z/SO(3) setting. G-S represents LRF built from Gram-Schmidt process.
}
\label{tab:ptb}
\end{center}
\end{table}
\subsection{Robustness to the Number of Neighbours}
The number of neighbors in KNN operation determines the scope of local feature extraction. We conduct experiments to investigate the robustness of LocoTrans to the number of neighbors. As shown in Tab.~\ref{tab:k}, our method achieves the best performance with $K=20$.
\begin{table}[htbp]
\begin{center}
\begin{tabular}{cccc}
\hline
\multicolumn{1}{c|}{Number of neighbors}    & $K=10$                   & $K=20$                   & $K=40$                   \\ \hline
\multicolumn{1}{c|}{Ours} & 91.3                 & 91.6                 & 90.8                 \\ \hline
\multicolumn{1}{l}{}  
\end{tabular}
\vspace{-15pt}
\caption{Classification accuracy (\%) under different neighbor size $K$ on ModelNet40 z/SO(3) setting. }
\vspace{-25pt}
\label{tab:k}
\end{center}
\end{table}

\subsection{More Backbones}
We further conduct experiments to investigate the robustness of other mainstream point cloud analysis models to rotations and explore the effects of LocoTrans on these backbones.
In Tab.~\ref{tab:bb}, we introduce three networks: 1) PointNet++~\cite{NIPS2017_d8bf84be}, a classic framework using PointNet~\cite{Qi_2017_CVPR} as local extractors, 2) the attention-based method PCT~\cite{Guo_2021}, which captures relationships between points well, and 3) PointMLP~\cite{ma2022rethinking}, which gives a pure residual MLP to replace sophisticated local extractors.
Although these backbones can achieve state-of-the-art on the well-aligned data, they fail to accurately classify the rotated data in z/SO(3) setting. 
In contrast, their performance is significantly improved when using our LocoTrans.
The results show that mainstream point cloud models lack rotation robustness, and LocoTrans can effectively address this issue.
\begin{table}[htbp]
\small
\begin{center}
\begin{tabular}{c|ccc}
\hline
\multirow{2}{*}{Row} & \multirow{2}{*}{Backbone} & \multicolumn{2}{c}{z/SO(3)} \\ \cline{3-4} 
                     &                           & w/o LocoTrans & w LocoTrans \\ \hline
\#1                    & PointNet++~\cite{NIPS2017_d8bf84be}         & 28.6             & 90.7           \\
\#2                    & PCT~\cite{Guo_2021}         & 25.6             & 90.7           \\
\#3                    & PointMLP~\cite{ma2022rethinking}                  & 31.1             & 91.1        \\ \hline
\end{tabular}
\caption{Classification accuracy (\%) on ModelNet40 z/SO(3) setting. `w/o LocoTrans' denotes not using LocoTrans while `w LocoTrans' represents applying LocoTrans on these backbones.}
\label{tab:bb}
\end{center}
\end{table}
\subsection{Model Output}
\label{secsm:output}
Here we analyze three types of outputs from the invariant branch, the equivariant branch, and fusion in our network. From Tab.~\ref{tab:output}, we can see fusion can significantly improve performance under different datasets and different settings. 
In addition, for z/SO(3) and SO(3)/SO(3) settings, both invariant branch and equivariant branch achieve similar performance in ModelNet40 while they suffer from performance fluctuation in ScanObjectNN, especially the equivariant branch.
We guess the reason is that
we use the vector-based equivariant network~\cite{Deng_2021_ICCV} as our equivariant branch, which combines input vectors linearly to achieve equivariance and thus has limited learning ability. 
Hence, facing changes in randomness brought by different rotations in two settings, equivariant branch cannot achieve relatively stable and consistent performance in ScanObjectNN containing background noise.
\begin{table}[htbp]
\begin{center}
\resizebox{1.0\linewidth}{!}{
\begin{tabular}{ccccc}
\hline
\multirow{2}{*}{Source} & \multicolumn{2}{c}{ModelNet40} & \multicolumn{2}{c}{ScanObjectNN} \\ \cline{2-5} 
                        & z/SO(3)      & SO(3)/SO(3)     & z/SO(3)       & SO(3)/SO(3)      \\ \hline
invariant branch        & 90.6         & 90.4            & 82.3          & 83.1             \\
equivariant branch      & 90.2         & 90.3            & 79.2          & 76.9             \\
fusion                  & 91.6         & 91.5            & 85.0          & 84.5             \\ \hline
\end{tabular}}
\caption{Classification accuracy (\%) on ModelNet40 dataset and ScanObjectNN dataset.}
\label{tab:output}
\end{center}
\end{table}

\section{Limitation}
\label{secsm:limit}
Although efficient, our approach may encounter a potential challenge when handling real-world data containing background noise.  
LocoTrans relies on equivariant features and as a result, its performance might be influenced by the efficacy of the equivariant branch. As mentioned in Sec.~\ref{secsm:output}, due to limited learning ability and changes in randomness, our equivariant branch suffers from performance fluctuation under z/SO(3) and SO(3)/SO(3) settings in ScanObjectNN, hindering our network from yielding better performance on both settings. 
In future work, we aim to improve the equivariant branch to address this issue. 

\section{Visualization}
\label{secsm:vis}
\begin{figure*}[htbp]
  \centering
  \includegraphics[width=0.7\linewidth]{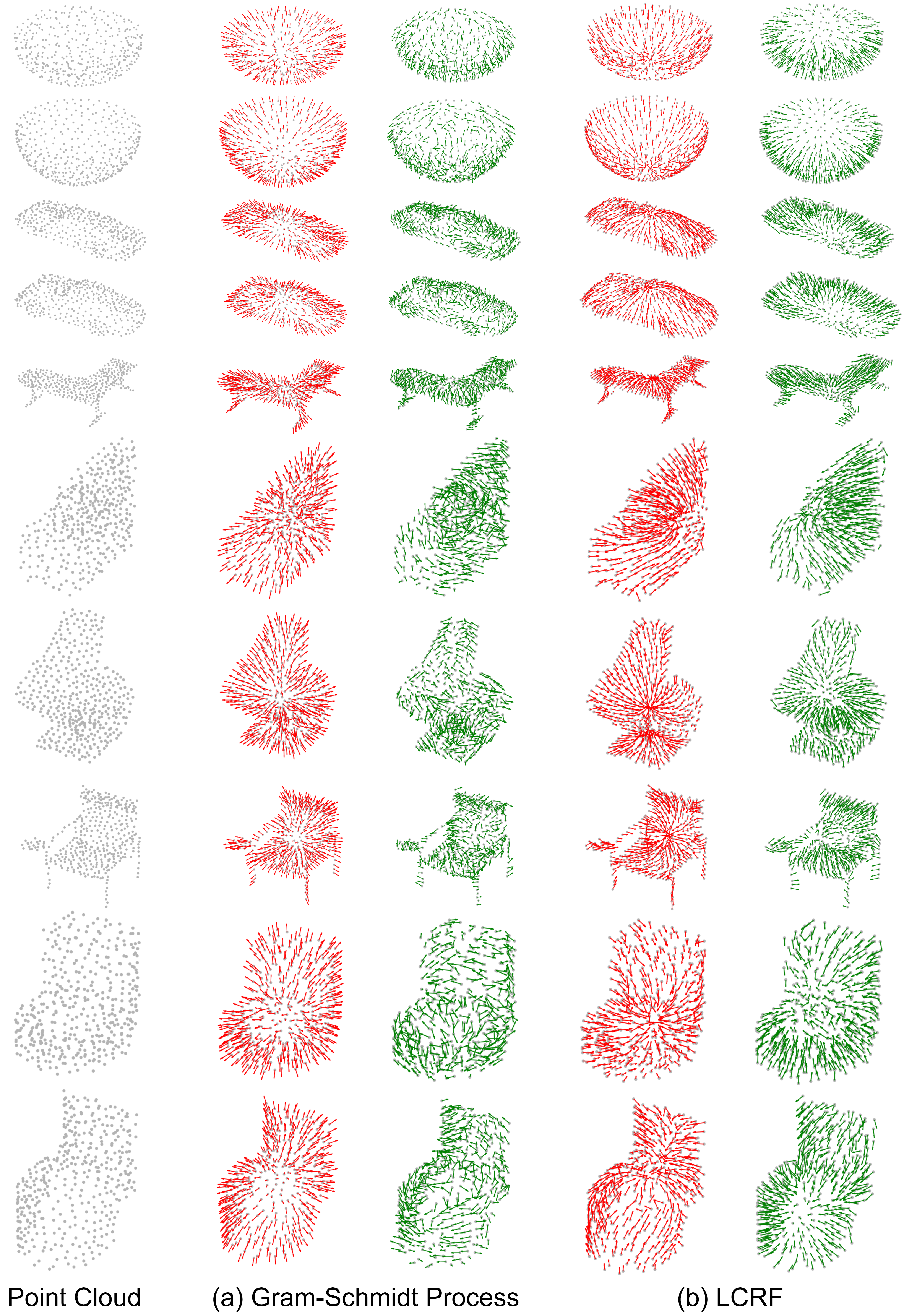} 
      \caption{
      Visualization of the learned orientations for the LRF built by the Gram-Schmidt process and our LCRF. 
      }
      \label{fig:vis}
\end{figure*}

Here, we provide additional visualization results of LRF built by the Gram-Schmidt process and our LCRF in Fig.~\ref{fig:vis}(a) and Fig.~\ref{fig:vis}(b) separately. The results demonstrate that our method achieves local consistency along different axes (Red and Green), while the previous LRF only works effectively along one axis (Red).

\section{Post-CVPR}
Considering the additional computational burden introduced by the equivariant branch affects the efficiency of our network, to address it, we attempt to decrease the complexity of the equivariant branch.
The experimental results are reported in Tab.~\ref{tab:complexity_new}. From the comparison between `Ours' and `Ours*' in Tab.~\ref{tab:complexity_new}, we can see reducing complexity does not have a significant negative impact on the performance of our method, proving that the effectiveness of our method does not solely depend on the increase of parameters and FLOPs.  Moreover, compared to  ~\citet{Li_2021_ICCV} and PaRI-Conv~\cite{Chen_2022_CVPR}, `Ours*' achieves a better trade-off between efficiency and performance.
\begin{table}[h]
\begin{center}
\scalebox{0.7}{
\begin{tabular}{c|cccc}
\hline
         &  ~\citet{Li_2021_ICCV}   & PaRI-Conv~\cite{Chen_2022_CVPR} &  Ours & Ours*  \\ \hline
Params   &  2.91M & 1.85M    & 6.27M         & 2.29M\\
FLOPs    &  3747M & 1938M    & 7998M         & 3351M \\
Acc.(ModelNet40) & 90.2  & 91.4     & 91.6          & 91.5 \\
Acc.(ScanObjectNN) &  84.3  & 77.8     & 85.0          & 84.9 \\\hline
\end{tabular}
}
\end{center}
\vspace{-15pt}
\caption{Computational burden and classification accuracy (\%) on ModelNet40 dataset and ScanObjectNN dataset, in z/SO(3) setting. `Ours*' reduces the complexity of equivariant branch compared to `Ours'. }
\label{tab:complexity_new}
\end{table}

\end{document}